\documentclass[conference]{IEEEtran}
\IEEEoverridecommandlockouts
\usepackage{cite}
\usepackage{amsmath,amssymb,amsfonts}
\usepackage{algorithmic}
\usepackage{graphicx}
\usepackage{textcomp}
\usepackage{booktabs}
\usepackage[switch]{lineno}
\usepackage{arydshln}
\usepackage{xcolor}
\def\BibTeX{{\rm B\kern-.05em{\sc i\kern-.025em b}\kern-.08em
    T\kern-.1667em\lower.7ex\hbox{E}\kern-.125emX}}
\begin{document}
\title{Fair-Net: A Network Architecture For Reducing Performance Disparity Between Identifiable Sub-Populations \\
}

\author{
  Arghya Datta$^{1}$,
  S. Joshua Swamidass$^{2,\dagger}$\\

  $^1$Department of Computer Science and Engineering, Washington University in Saint Louis\\
  $^2$Department of Pathology and Immunology, Washington University in Saint Louis\\
  $^\dagger$swamidass@wustl.edu
}

\maketitle

\begin{abstract}

In real world datasets, particular groups are under-represented, much rarer than others, and machine learning classifiers will often preform worse on under-represented populations. This problem is aggravated across many domains where datasets are  class imbalanced, with a minority class far rarer than the majority class. Naive approaches to handle under-representation and class imbalance include training sub-population specific classifiers that handle class imbalance or training a global classifier that overlooks sub-population disparities and aims to achieve high overall accuracy by handling class imbalance. In this study, we  find that these  approaches are  vulnerable in class imbalanced datasets with minority sub-populations. We introduced Fair-Net, a branched multitask neural network architecture that improves both classification accuracy and probability calibration across identifiable sub-populations in class imbalanced datasets. Fair-Nets is a straightforward extension to the output layer and error function of a network, so can be incorporated in far more complex architectures. Empirical studies with three real world benchmark datasets demonstrate that Fair-Net improves classification and calibration performance, substantially reducing performance disparity between gender and racial sub-populations. 
\end{abstract}


\section{Introduction}

Decision-making systems, based on neural network architectures, are widely used in many critical tasks such as criminal justice~\cite{doi:10.1177/1477370819876762}, granting loans~\cite{doi:10.1098/rsos.191649}, skin cancer detection~\cite{Esteva2017} and face recognition~\cite{merler2019diversity}. However, there have been growing concerns regarding the performance disparities of these decision making systems across many sensitive domains where there are under-represented sub-populations in the underlying training dataset or application domain.

Under-representation is when samples from a particular sub-population such as those based on gender or race are rare. Under these circumstances, classifiers tend to exhibit disparate performance, with greater accuracy on the majority sub-populations than the minority sub-populations. A previous case study by Buolamwini et al.~\cite{pmlr-v81-buolamwini18a} known as "Gender Shades" highlighted noticeable performance disparities in facial recognition systems between male and female sub-populations for classification tasks. Even though these classification systems achieved more than 90\% global classification accuracy for gender detection, classification accuracy was much higher in light skinned individuals than dark-skinned ones. 


Numerous studies have been conducted that highlight the problem of under-representation in datasets~\cite{Kearns2019AnES,DBLP:conf/aies/KimGZ19,10.5555/3157382.3157469,8452913}. The problem of under-representation is even more challenging in presence of class imbalance. Class imbalance occurs when samples from one class are far more rare than the other. Classifiers are usually biased towards the majority class, thereby performing poorly on the minority class. Since machine learning classifiers are commonly used in decision-making systems, they should simultaneously be accurate as well as produce well calibrated probabilities. Predictions from a binary classifier are said to be well-calibrated if the outcomes predicted to occur with a probability \textit{p} occur \textit{p} fraction of the time. Since classifiers minimize error on training, it is often a common practise to assign high costs on misclassifications on the minority class so as to maximize the classification performance on the minority class but overlooking the calibration performance. Common parametric and non-parametric approaches such as Platt scaling~\cite{Platt99}, isotonic regression~\cite{Zadrozny2002IsoReg} and Bayesian binning into quantiles (BBQ)~\cite{Naeini2015} are often used to post-process classifiers' outputs for probability calibration. However, previous research~\cite{huang2020} has shown that the common parametric and non-parametric calibration techniques are often unstable on class imbalanced datasets. Even though, a classifier is trained and post-processed to maximize classification and calibration accuracy for class imbalanced datasets, there may still be significant performance drops across under-represented sub-populations present in the dataset population.

Little work has been done to develop neural network architecture that jointly learns classification and calibration in under-represented sub-populations while handling the skewed distribution of the minority and majority samples in class imbalanced datasets. Recently, the Cal-Net neural network architecture~\cite{cal-net_cite} demonstrated simultaneous improvement in classification and calibration performance on class imbalanced datasets. Here, we aim to build on this  architecture to improve predictive performance across multiple sub-populations with Fair-Net: a neural network architecture that simultaneously optimizes classification and calibration performances across identifiable sub-populations in the dataset population. Empirically, we find that Fair-Net achieves the best classification and calibration performances across diverse sub-populations of interest.

\section{Related Works and Methodologies}

Prior research works and methodologies that have been proposed to handle class imbalance and probability calibration while improving classification performance across under-represented sub-populations in datasets.

Previous research studies have proposed parametric and non-parametric post-processing probability calibration techniques such as Platt scaling~\cite{Platt99}, isotonic regression~\cite{Zadrozny2002IsoReg}, histogram binning~\cite{Zadrozny2001Binning} and bayesian binning into quantiles (BBQ)~\cite{Naeini2015}. The post-processing calibration techniques utilize a holdout validation dataset for re-scaling the base classifiers' outputs to improve calibration performance thereby reducing the effective number of sampling for training the base classifier. In datasets where the number of samples is low, this may often lead to under-trained classifiers.

Class imbalance is a widespread challenge in machine learning and previous studies have proposed several strategies to mitigate this problem. Sampling is a common approach to mitigate class imbalance. Common sampling strategies include over-sampling~\cite{Ling98}, where samples from the minority class is re-sampled randomly to eliminate the skewness from the data distribution. Similarly, under-sampling~\cite{Kubat97} eliminates samples from the majority class randomly to match the distribution of the minority class. Methods such as 
synthetic minority over-sampling technique (SMOTE)~\cite{Chawla2002} has been proposed that removes the skewness from the imbalanced data distributions by  generating synthetic minority class samples. Cost-sensitive learning~\cite{Domingos1999,Elkan2001} and sample weighting\cite{ting1998} are commonly used to assign high weights to samples from the minority class by modifying the objective function. Even though sampling strategies are widely used for managing class imbalance, there are well-known pitfalls such as overfitting~\cite{Holte1989} due to over-sampling as well as information loss ~\cite{Tang2009} and inducing bias in calibration due to under-sampling~\cite{DalPozzolo2015}.

A naive approach to address challenges in predictive modeling across sub-populations of interest in a dataset is to train a separate classifier on each sub-population of interest while simultaneously using previously proposed strategies to handle class imbalance. We have included this approach as a baseline in our study. We find that this approach performs poorly in minority sub-populations where only a small number of samples are available to train sub-population specific classifiers. To overcome this shortcoming, branched neural network architectures can be used where each branch is trained on different sub-populations so as to improve the predictive performance for that specific sub-population. In our ablation studies, we have showed that this approach does not  simultaneously achieve the best classification and calibration performances in minority sub-populations in class imbalanced datasets. Previous research studies have proposed methodologies~\cite{Kearns2019AnES,DBLP:conf/aies/KimGZ19} to improve classification accuracy across sub-populations as well as various definitions of fairness such as equalized odds and equal opportunity~\cite{10.5555/3157382.3157469}, demographic parity~\cite{8452913} etc. Our definition of fairness is different from parity based notions~\cite{10.5555/3157382.3157469,8452913}. Instead, we increase fairness by reducing disparity in classification and calibration performance across sub-populations. Disparity is defined as the variance of performance across identifiable sub-populations of interest in class imbalanced datasets.

\begin{figure}[htb!]
 \centering
  \includegraphics[width=\columnwidth]{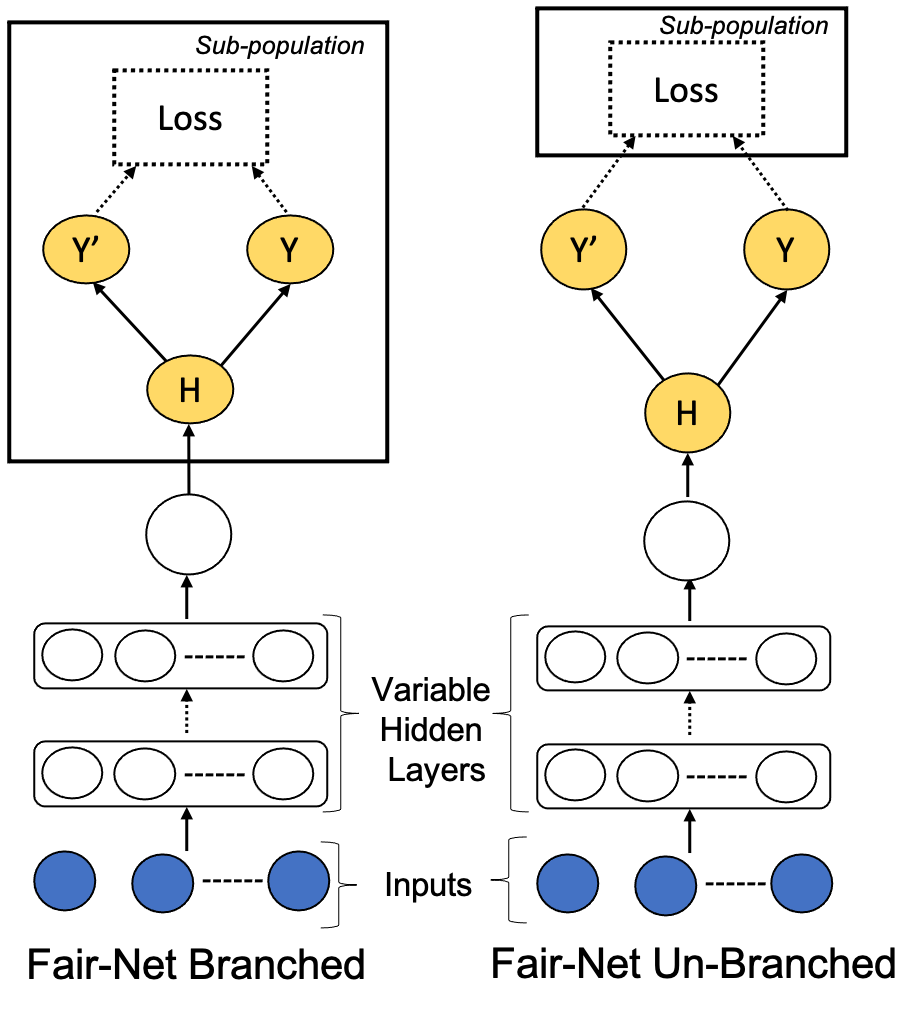}
    \caption{Different variants of Fair-Nets. Fair-Net Branched trains a different branch for each sub-population, whereas Fair-Net Un-Branched computes the total loss by summing over losses for each sub-population of interest.}
\label{fig:Fair-Net_models}
\end{figure}

\section{Materials and Methods}

\subsection{The Fair-Net Architecture}

The Fair-Net architecture expands the  Cal-Net architecture, which developed to improve calibration on imbalanced datasets~\cite{cal-net_cite}. Like the Cal-Net architecture, the Fair-Net architecture transforms the binary classification problem into a multi-task problem using two outputs (Figure~\ref{fig:Fair-Net_models}). The primary output ($Y$) is tuned to produce well-scaled probabilities, whereas the secondary output ($Y'$) is utilized only during the training phase to maximize the classification performance by upweighting samples from the minority class to be equally prevalent as samples from the majority class. 

Mirroring  Cal-Net , the primary ($Y$) and secondary ($Y'$) outputs in Fair-Nets, computed using logistic activation functions, are computed from a hidden layer ($H$) with a single node. This structure ensures that the neural network architecture enforces a monotonic relationship between the primary and the secondary outputs. Both the outputs are monotonic functions of a single number $H$ so they are monotonic transformations of each other. 

We have introduced two broad variants of Fair-Nets: ``Fair-Net Branched" and ``Fair-Net Un-Branched". In the first variant, "Fair-Net Branched" trains a different ``branch" consisting of a primary output ($Y$) and a secondary output ($Y'$), computed using a single hidden node $H$, for each sub-population of interest. The primary output ($Y$) in a ``branch" is tuned to produce well-calibrated probabilities, whereas the secondary output ($Y'$) is tuned to maximize classification performance by upweighting samples from the minority class in the sub-population to be equally prevalent as samples from the majority class for that sub-population. This modification requires six trainable parameters in total, with three weights and three biases for each sub-population of interest. Thus, each branch is tuned to maximize the classification and calibration performances for each sub-population of interest at the cost of more trainable parameters. 
 
In the second variant, ``Fair-Net Un-Branched," a single ``branch" consisting of a primary output ($Y$) and a secondary output ($Y'$), computed from a single hidden node $H$, is tuned to maximize the classification and calibration performance across all the sub-populations of interest. Unlike the variant ``Fair-Net Branched", the variant ``Fair-Net Un-Branched" does not result in additional trainable parameters for each sub-population of interest. 

\subsection{Loss Components}

All the variants of Fair-Nets make use of the same loss components used by Cal-Net~\cite{cal-net_cite}. 

The primary output, $Y=\{y_{g,i}\}$, indexed by instance, $i$, in sub-population, $g$, for both ``Fair-Net Branched" and ``Fair-Net Un-Branched" utilize a logistic activation function. The loss component for each sub-population, $g$, are computed based on this output and the target class labels $T=\{t_{g,i}\}$.

The first loss component, $L_{X,g}$, is the binary cross entropy error between $Y$ and $T$ for sub-population $g$. The instances in the majority class for each sub-population contribute more to the loss in class imbalanced datasets.

The second loss component, $L_{B,g}$, computes the balanced cross-entropy loss for each sub-population, $g$, between $T$ and $Y'$. Instances from the minority class for each sub-population is upweighted to be equally prevalent as samples from the majority class for that sub-population. In all the variants of Fair-Nets, the majority (negatives) class samples in sub-population $g$ are weighted as $N_g/2n_g$ and the minority (positives) class samples are weighted as $N_g/2p_g$ where $N_g$ is the number of samples in sub-population $g$ and $p_g$ and $n_g$ are the number of samples in the minority (positives) and majority (negatives) classes, respectively. This weighting scheme ensures that for each sub-population $g$, the minority and majority classes are weighed equally.

The total loss function ($L$) for Fair-Net Branched and Fair-Net Un-Branched is computed as,
\begin{equation}
L =\sum_{g\in G} \lambda_g \cdot [L_{X,g} + L_{B,g}],
\end{equation}
where $G$ is the set of all sub-populations of interest and $\lambda_g$ is a hyper-parameter that can be tuned to assign higher mis-classification costs for sub-population $g$. In all our experiments,  $\lambda_g=1$ for all sub-populations.

We also use the histogram loss from Cal-Net ~\cite{cal-net_cite} on the primary output $Y$ for generating well scaled-probabilities. In a well-calibrated probabilistic model for binary classification tasks, the proportion of positive examples in each bin of a reliability diagram should match the average of the predictions for the bin, which is usually close to the midpoint of the bin. Hence, the histogram loss, $L_{H,g}$ for each sub-population $g$ is computed as the RMSE between the proportion positives and the midpoints of the bin. 

The total loss function ($L$) for Fair-Net Branched [histogram loss] and Fair-Net Un-Branched [histogram loss] is computed as,
\begin{equation}
L =\sum_{g\in G} \lambda_g \cdot [L_{X,g} + L_{B,g} + \lambda_{H,g} L_{H,g}]
\end{equation}
where $L_{H,g}$ is the histogram loss for sub-population $g$ and $\lambda_{H,g}$ is a hyper-parameter that can be used to tune $L_{H,g}$. Other formulations of the histogram loss may be effective, but exploring them is left for future work. 
Empirical analyses show that all the loss components are necessary to optimize classification and calibration across under-represented sub-populations in class imbalanced scenarios. 

\subsection{Datasets}
For our experiments, we used three datasets namely (Table~\ref{table:benchmark_stats}): (1) Propublica COMPAS dataset~\cite{propublica_compas_dataset} (2) UCI credit card default dataset~\cite{YEH20092473} and (3) UCI adult census dataset~\cite{Kohavi}.
\begin{itemize}
    \item Propublica COMPAS dataset: We used a smaller subset of the propublica COMPAS dataset~\cite{propublica_compas_dataset} consisting of 6172 instances with 5 features. A binary target variable indicated if an individual would re-offend within the next two years. We used the gender variable to consider two sub-populations namely: Female (F) and Not Female (NF). The imbalance ratios of the target variable in F and NF sub-populations were 1.8 and 1.08 respectively. 
    \item UCI credit card default dataset: The credit card default dataset~\cite{YEH20092473} from UCI~\cite{uci_repository} repository consists of 30,000 instances with 23 features. The binary target variable indicated whether an individual would incur a default payment or not. We have considered two sub-populations based on gender namely: Male (M) and Female(F). The imbalance ratios for the target variable for M and F sub-populations are 3.14 and 3.81 respectively.
    \item UCI Adult Census dataset: The adult census dataset~\cite{Kohavi} from UCI~\cite{uci_repository} repository consists of 48,842 instances with 14 features and a binary target variable that indicated if an individual earned more than \$$50,000$ or not. After removing samples with missing values, 45,222 samples were used for the analyses. We have considered 8 sub-populations based on gender and race namely: Male (M), Female (F), Black (B), White (W), Black Male (BM), Black Female (BF), White Male (WM) and White Female (WF). 
\end{itemize}

\subsection{Training and Evaluation Protocol}

We evaluated the variants of Fair-Nets and baseline models for probability calibration performance and classification accuracy using a stratified train, validation and test split. For each dataset, we kept a stratified split of the dataset as a test set($20-25 \%$ of the dataset) such that the percentages of sub-populations and the imbalance ratios for each sub-population are preserved across train, validation and test sets. Since most of these datasets have a low number of samples for the minority sub-populations, this strategy ensured that sufficient minority samples are present in all the splits. For our experiments, we trained the variants of Fair-Nets with a single hidden layer of 5 units with exponential linear unit (ELU) activation and L2 regularization.

\subsection{Baselines}
As a baseline for comparison with different variants of Fair-Nets, we trained neural network (NN) architectures with one ELU activated hidden layer consisting of 10 hidden units. We used balanced cross entropy loss to train these neural network architectures such that the samples from the minority class are upweighted to be equally prevalent as samples from the majority class. Balanced cross entropy loss usually improves the classification performance in class imbalanced datasets~\cite{Johnson2019}. Also, we trained neural network architectures using the same architecture (one hidden layer with 10 ELU activated hidden units) for each sub-population of interest using balanced cross-entropy losses. Finally, we trained Cal-Net architectures with one hidden layer consisting of 5 hidden units with ELU activation. For our case studies, all the variants of Fair-Nets usually had less number of trainable parameters than the baseline NN [balanced xent] and sub-population specific baselines.

\subsection{Assessment Metrics}
We evaluate the predictive performance of Fair-Nets and the associated baselines on different sub-populations by reporting the maximum F-measure and area under the receiver operating characteristic (ROC AUC). Previous research work~\cite{Jesse2006} has shown that ROC AUC is often unreliable in class imbalanced datasets. On the contrary, F-measure is a commonly used metric to summarize classification performance in class imbalanced datasets. We highlight the imbalance ratio (IR), calculated as $\frac{n_0}{n_1}$, where $n_1$ is the number of minority (positives) samples and $n_0$ is the number of majority (negatives) samples across different sub-populations of interest in the datasets. In order to summarize the calibration performance of Fair-Nets and the associated baselines, we have reported the expected calibration error (ECE)~\cite{Naeini2015,Kueppers_2020_CVPR_Workshops} and utilized reliability diagrams~\cite{degroot_reliability}. A classifier that achieves higher F-measure and higher AUROC along with lower ECE across different sub-populations of interest is preferred.

\section{Results \& Discussion }

\begin{table}[!htb]
\centering
\caption{Class Imbalance statistics for real world datasets}
\begin{tabular}{lcccc}
\toprule
Dataset                     & Size       & \% +ve    & IR    & No. of test samples\\\toprule
COMPAS                      & 6172       & 45.5      & 1.2   & 1237                \\
Credit default              & 30000      & 22.12     & 3.52  & 7502               \\
Census income               & 45222      & 24.78     & 3     & 11130              \\\bottomrule
\end{tabular}
\label{table:benchmark_stats}
\end{table}

\begin{figure*}[htb!]
 \centering
  \includegraphics[width=\textwidth]{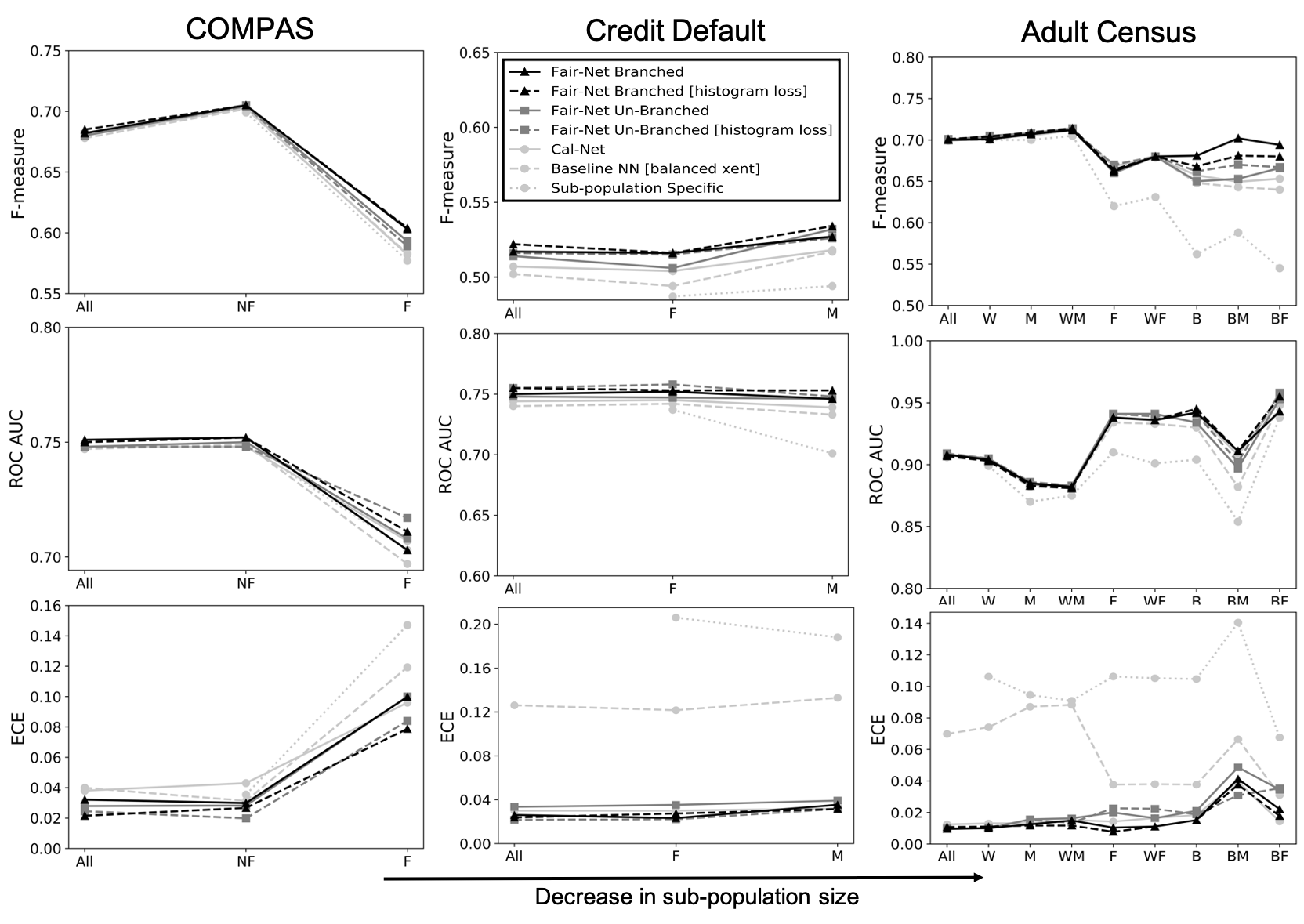}
    \caption{On the benchmark datasets, variants of Fair-Net achieved the best predictive performance in terms of F-measure, ROC-AUC and ECE across different sub-populations of interest.}
\label{fig:benchmark_results}
\end{figure*}

\subsection{Propublica COMPAS dataset}
Fair-Net variants outperformed the baselines on the Propublica COMPAS dataset in classification and calibration performances both on the overall population as well as on the different sub-populations of interest (Figure~\ref{fig:benchmark_results}). For the F sub-population, the IR for the target variable was 1.8 , which was higher than that of the overall sub-population. Fair-Net variants outperformed the baselines on the F sub-population in predictive performance by achieving the highest F-measure while simultaneously achieving the lowest ECE, thereby improving the calibration performance. Both the variants of Fair-Net Branched outperformed the variants of Fair-Net Un-Branched in classification performance owing to a greater number of available trainable parameters. For both Fair-Nets Branched and Fair-Nets Un-Branched variants, the inclusion of the histogram loss helped in improving calibration performance by reducing ECE. This highlights the potential benefit of incorporating histogram loss to further improve the calibration performance across sub-populations. Sub-population specific baseline models, trained exclusively for each sub-population of interest, usually exhibited poor predictive performance due to the availability of a lower number of samples in the training dataset. As empirically shown, all the variants of Fair-Nets improved predictive performance in the under-represented F sub-population with a high IR than the overall population.

\subsection{UCI Credit card default dataset}

All the variants of Fair-Nets outperformed the baselines in classification and calibration performance on the overall population while simultaneously improving predictive performances across M and F  sub-populations (Figure~\ref{fig:benchmark_results}). We observed similar trends in this case study as well where both the variants of Fair-Net Branched outperformed the variants of Fair-Net Un-Branched at the cost of more trainable parameters. Both the variants of Fair-Nets with histogram loss outperformed the corresponding variants without histogram loss in calibration performance by achieving lower ECE scores. All the variants of Fair-Nets improved the predictive performance in the under-represented M sub-population in the dataset. 

\subsection{UCI Adult Census dataset}
\begin{figure*}[htb!]
 \centering
  \includegraphics[width=\textwidth]{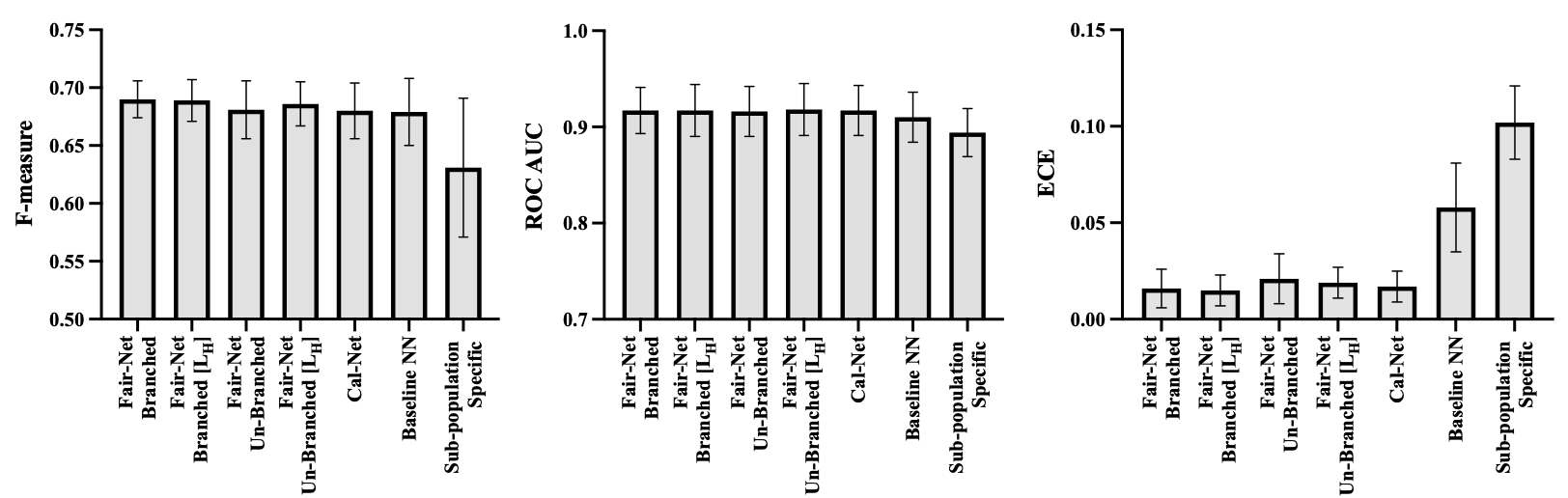}
    \caption{Variants of Fair-Net achieved the best predictive performance (high F-measure and ROC AUC alongside low ECE) with lowest variance across identifiable sub-populations of interest for the adult census dataset~\cite{Kohavi}. $L_H$ refers to the histogram loss.}
\label{fig:variations}
\end{figure*}

\begin{figure*}[htb!]
 \centering
  \includegraphics[width=\textwidth]{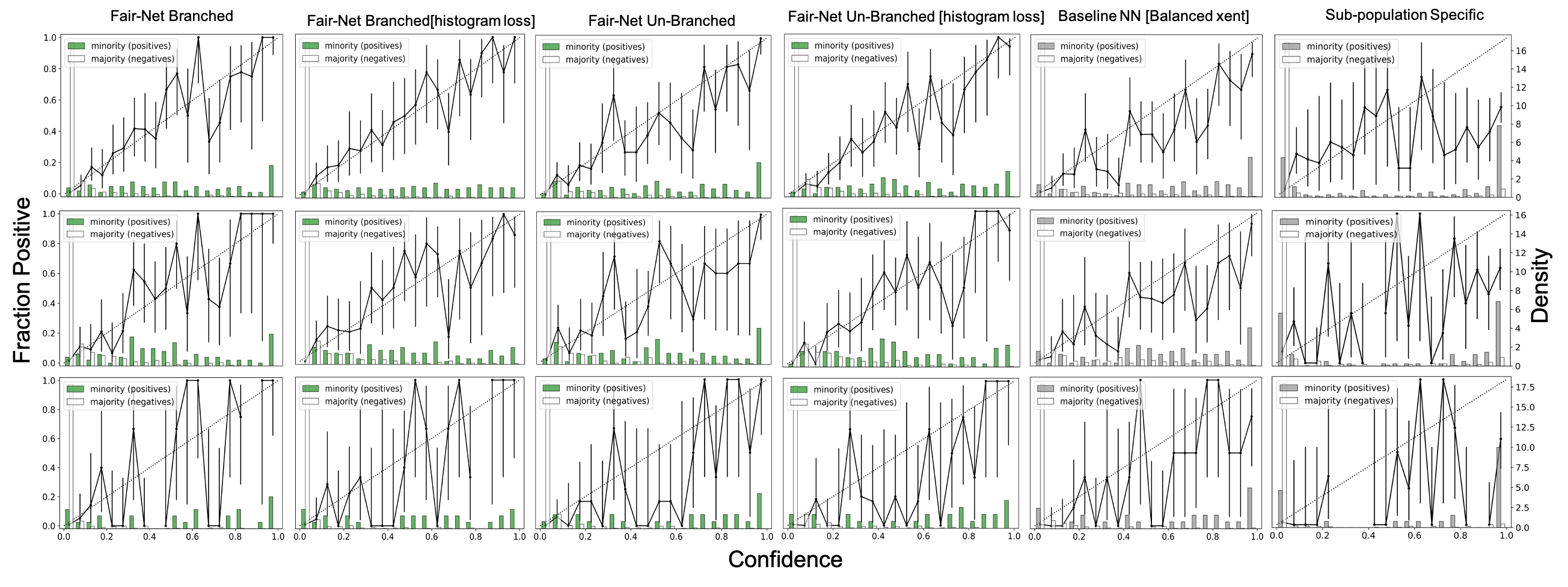}
    \caption{ Reliability diagrams on the adult census dataset (bins=20) showing that Fair-Net variants have the best calibration performance. Top row: Black (B) sub-population; Middle row: Black Male (BM) sub-population; Bottom row: Black Female (BF) sub-population}
\label{fig:census_reliability}
\end{figure*}

Similar to the prior two case studies, variants of Fair-Nets achieved the best classification and calibration performance across different sub-populations in the adult census dataset~\cite{Kohavi}. Eight sub-populations were considered in this case study: Male (M), Female (F), Black (B), White (W), Black Male (BM), Black Female (BF), White Male (WM) and White Female (WF). Out of all these sub-populations, B, BM and BF sub-populations were the most under-represented, accounting for less than 10\% of the overall population. Furthermore, the IR values across B, BM and BF sub-populations are 6.8, 4.26 and 14.7 respectively. Empirical results (Figure~\ref{fig:benchmark_results}) showed that the variants of Fair-Nets outperformed the baselines across all the eight sub-populations as well as on the overall population in classification and calibration performances (Figure~\ref{fig:census_reliability}). Moreover, the improvements in classification performance for the variants of Fair-Nets were noticeable in the B, BM and BF subpopulations, where, all the variants of Fair-Nets outperformed the baselines by achieving higher F-measure and ROC AUC and lower ECE. The baseline neural network, trained using balanced cross entropy loss for the overall population, achieved similar classification performance to Fair-Nets in the majority sub-populations but incurred drops in F-measure in the under-represented sub-populations such as B, BM and BF. The sub-population specific baseline models, trained exclusively on different sub-populations, performed poorly in the under-represented sub-populations with high class imbalances owing to a shortage in training samples.  On the contrary, branched variants of Fair-Net achieved significantly higher predictive performance by adding six trainable parameters (three weight variables and three bias variables) for each sub-population of interest, whereas the un-branched variants of Fair-Nets did not add any additional trainable parameters for different sub-populations of interest. The branched variants of Fair-Net outperformed the un-branched variants in predictive performance due to the availability of more trainable parameters in the neural network architecture. Furthermore, the average classification and calibration performances of Fair-Net variants across different sub-populations are higher than the baselines with low standard deviation (Figure~\ref{fig:variations}). This suggests that Fair-Nets do not incur any substantial performance drops in under-represented sub-populations with high class imbalance ratios. Finally, the classification error (\%) across different sub-populations for the Fair-Net variants are comparable with prior published work by Kim et al.~\cite{DBLP:conf/aies/KimGZ19}. We observed that Fair-Net Branched achieved lower classification error (Table~\ref{tab:compare_prior_pubs}), suggesting that Fair-Nets achieve the best predictive accuracy in the under-represented sub-populations for the adult census dataset. However, we note that classification error(\%) is a poor metric in class imbalanced sub-populations since a classifier with low classification error (\%) may not achieve high F-measure and performs substantially worse in class imbalanced datasets~\cite{LUQUE2019216}.

\begin{table*}[!htb]
\centering
\caption{Comparing performance of Fair-Net with published results in the literature such as Multi-accuracy~\cite{DBLP:conf/aies/KimGZ19} on similar, but not identical, test dataset.}
\begin{tabular}{cccccccccc}
\toprule
\multicolumn{10}{c}{\textbf{Classification Error (\%) $\downarrow$}} \\  \midrule

                                               & all             & W               & M              & WM 	          & F              & WF 		      & B                 & BM              & BF                        \\
									           \cmidrule(r){2-10}

Fair-Net Branched                               & 16.29	         & 17.09	       & 20.1	        & 21.7	          & 7.67	       & 7.95	          & 9.5 	          & 13.25	        & 3.83        \\
\cmidrule(r){2-10}
Multi-Accuracy~\cite{DBLP:conf/aies/KimGZ19}   & 14.7	         & 15	           & 18.3	        & 18.3	          & 7.2	           & 7.3	          & 9.4	              & 13.9	        & 4.5         \\\bottomrule
\end{tabular}	
\label{tab:compare_prior_pubs}
\end{table*}

\begin{figure}[!htbp]
 \centering
  \includegraphics[width=7.4cm,height=22cm]{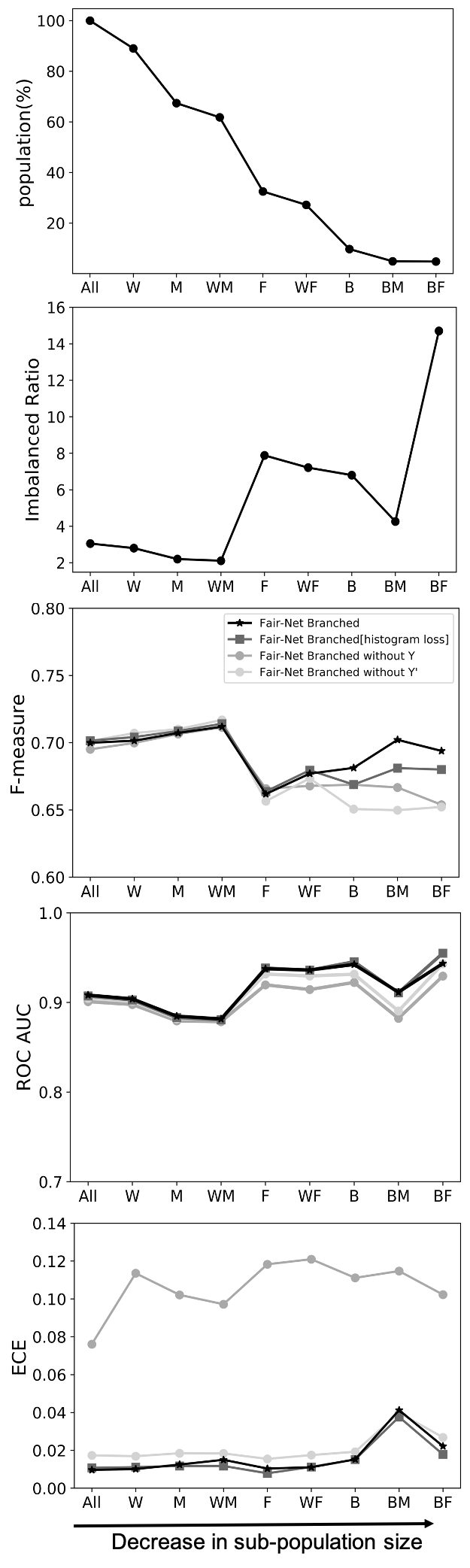}
    \caption{Ablation analyses demonstrating different components of Fair-Net are essential to improve classification and calibration performances for the adult census income dataset~\cite{Kohavi}.}
\label{fig:census_ablation}
\end{figure}

\subsection{Ablation analyses}
Ablation analyses demonstrated the importance of different components in the Fair-Net's architecture. We trained multi-task neural network architectures that resembled variants of Fair-Net Branched by removing (1) the primary output $Y$ and (2) the secondary output $Y'$ from each of the sub-population heads in Fair-Net's architecture to evaluate whether the classification and calibration performances are affected. For the ablation analyses, we focused on the adult census dataset~\cite{Kohavi} since it contains sub-populations with a diverse range of samples and imbalance ratios.

\subsubsection{With and without primary output $Y$ for each sub-population}
We trained a modified architecture without the primary output $Y$ from every sub-population network head. Thus, the Fair-Net architecture was reduced to a branched neural network architecture, where each branch was trained on a different sub-population using a balanced cross-entropy loss. The balanced cross-entropy loss upweighted minority class samples to be equally prevalent as samples from the majority class for each sub-population. Empirical results (Figure~\ref{fig:census_ablation}) showed that this modified architecture incurred a drop in classification performance across minority sub-populations such B, BM and BF. Moreover, we observed that this architecture had poor calibration performance (Figure~\ref{fig:census_ablation}) when compared to a standard Fair-Net architecture with a primary output $Y$ for each sub-population. Post-processing this modified architecture's outputs using parametric and non-parametric calibration techniques may improve the calibration performance. The standard Fair-Net Branched architecture continued to achieve the best overall classification and calibration performances across all the sub-populations of interest.

\subsubsection{With and without secondary output $Y'$ for each sub-population}

We trained a modified architecture after eliminating the secondary output $Y'$ from each of the sub-population branches. This essentially reduced the Fair-Net architecture to a branched neural network architecture, where a separate network head was trained for each sub-population using non-weighted cross-entropy loss. Hence, samples from both the majority (negatives) and the minority (positives) classes were weighted equally. The multi-task architecture without the secondary output $Y'$ for each sub-population produced well calibrated probabilities. However, it achieved the lowest classification performance across all the under-represented sub-populations. In the standard Fair-Net architecture, the secondary output $Y'$ was trained using a balanced cross-entropy loss so that samples from the minority class were upweighted to be equally prevalent as samples from the majority class belonging to the same sub-population thereby improving classification performance. We hypothesized that this modified architecture without the secondary output $Y'$ would be comparatively weaker than the standard Fair-Net architecture in classification performance. 

Empirically, we observed that there were drops in  classification performance of this modified architecture across B, BM and BF sub-populations in terms of F-measure and ROC-AUC (Figure~\ref{fig:census_ablation}). As evident from the distribution of these sub-populations in the dataset, B, BM and BF had very few samples with a high class imbalance ratio when compared to other sub-populations. Hence, we concluded that the secondary output $Y'$ is necessary to improve the classification performance across minority sub-populations with high class imbalance ratios. Both the variants of Fair-Net Branched continued to achieve the 
best classification and calibration performances across all the sub-populations of interest.

\subsubsection{With and without histogram loss ($L_H$)}

We introduced two variants of Fair-Nets that used the histogram loss~\cite{cal-net_cite}. Our case studies across COMPAS data, credit card default dataset and adult census dataset showed that the variants of Fair-Nets trained using histogram loss often outperformed the corresponding variants of Fair-Nets without the histogram loss by achieving lower ECE scores. This suggests that optimizing on the histogram loss may result in improved calibration performance. There may be other formulations for the histogram loss and fully exploring options is left for future studies.

\subsection{Study limitations \& future directions}

This modeling framework requires that all sub-populations are identifiable from the outset. We explicitly identify each sub-populations in training the model,  summing over the losses for each sub-population. Hence, our modeling framework needs access to the features that were used to identify these sub-populations of interest. In our study, the secondary output $Y'$ equally weighs the minority and the majority samples for each sub-population. However, upweighting the minority samples using a higher weight may yield better results in class imbalanced scenarios. In general, neural network architectures may often exhibit poor predictive performance and generalization due to unavailability in training data for under-represented sub-populations~\cite{cui2015}. As a result, variants of Fair-Net may exhibit degraded performance in the absence of enough training data. In our case studies, we weighted each sub-population of interest equally by setting $\lambda_g$ as 1 for all $g \in G$. However, upweighting under-represented or minority sub-populations may result in improved predictive performance in these subpopulations and exploring options is left for future work. Exploiting multi-task architectures may prove to be an effective way to improve predictive performance across these sub-populations, as evident in our study.


\section{Conclusion}
In this work, we have introduced Fair-Net, a class of neural network architectures that simultaneously improved classification and calibration performances across diverse sub-populations of interest in class imbalanced datasets. Empirically, we showed that the variants of Fair-Net outperformed commonly used neural network architectures by achieving higher F-measure, ROC-AUC and low ECE across different sub-populations of interest in three real world datasets: UCI Credit card default dataset, UCI Adult census dataset and Propublica COMPAS datasets. Due to its simplicity, Fair-Nets can readily be incorporated in complex network architectures as the final layer to improve predictive performance across sub-populations of interest.

\section{Acknowledgement}
This research was supported by the National Library Of Medicine of the National Institutes of Health under award numbers R01LM012222 and R01LM012482 and the National Institute of General Medical Sciences under award number R01GM140635.

\bibliographystyle{unsrt}
\bibliography{bibliography/references}
\end{document}